\documentclass{article}

\usepackage{amsmath}
\usepackage{amsfonts}
\usepackage[table]{xcolor}

\usepackage{microtype}
\usepackage{graphicx}
\usepackage{subfigure}
\usepackage{booktabs} 

\usepackage{hyperref}

\usepackage[accepted]{mlsys2024}

\mlsystitlerunning{YOURA}

\begin{document}
\pagestyle {plain}
\twocolumn[
\mlsystitle{\underline{Y}ou \underline{O}nly \underline{U}se \underline{R}eactive \underline{A}ttention Slice\\for Long Context Retrieval}



\mlsyssetsymbol{equal}{*}

\begin{mlsysauthorlist}
\mlsysauthor{Yun Joon Soh}{ucsd}
\mlsysauthor{Hanxian Huang}{ucsd}
\mlsysauthor{Yuandong Tian}{meta}
\mlsysauthor{Jishen Zhao}{ucsd}
\end{mlsysauthorlist}

\mlsysaffiliation{ucsd}{Department of Computer Science and Engineering, University of California, San Diego, San Diego, USA}
\mlsysaffiliation{meta}{Meta AI (FAIR), Menlo Park, California, USA. Meta-affiliated author served in an advisory role and all data processing occurred at the University of California, San Diego}

\mlsyscorrespondingauthor{Jishen Zhao}{jzhao@ucsd.edu}

\mlsyskeywords{Long Context Retrieval, RAG}

\vskip 0.3in

\begin{abstract}
Supporting longer context for Large Language Models (LLM) is a promising direction to advance LLMs.
As training a model for a longer context window is computationally expensive, many alternative solutions, such as Retrieval Augmented Generation (RAG), have been used.
However, most existing RAG methods adopt embedding-based retrieval that falls short on long contexts.

To address such challenges, we propose an attention-based retrieval technique, \underline{Y}ou \underline{O}nly \underline{U}se \underline{R}eactive \underline{A}ttention slice (YOURA).
YOURA leverages a novel retrieval heuristic called \textit{reaction score} to rank the relevance of each sentence in the input context with the query sentence.
Intuitively, we measure how the per-token attention score ``reacts'' to the query and greedily retrieves the most reactive sentences.
Internally, YOURA generates a token-indexed vector (called \textit{reaction vector}) for the whole input context.
To map each sentence to the token-indexed vector, we propose an Embedding-Agnostic Sentence Yield (EASY), a best-effort token wiggling algorithm.

We evaluate our retrieval technique on three open-source pre-trained LLM models across six LongBench QA datasets. Our technique achieves up to 30\% vLLM inference throughput improvement for serving long-context queries --- with a nearly identical quality score to the simple yet effective truncate-middle approach~\cite{bai2023longbench}.

\end{abstract}

]



\printAffiliationsAndNotice{}  

\section{Introduction}

Pre-trained Large Language Models (LLM) are widely explored for various Natural Language Processing (NLP) tasks.
To enhance the model for more complex tasks, researchers seek ways to extend the LLM's context window size
via fine-tuning the pre-trained model \cite{chen2023longlora, peft}, proposing advanced attention mechanisms \cite{xiao2024efficient, zhang2023h2o}, and improving Retrieval Augmented Generation (RAG).
However, fine-tuning requires additional computational power for each pre-trained model and is often targeted for specific tasks or datasets.
Advanced attention mechanisms conceptually extend the context window by selectively choosing a subsequence of tokens for attention but are susceptible to details due to the nature of dropping information at the token level.
RAG splits the contextual text into smaller chunks and retrieves a subset of chunks before inferencing based on the semantical distance to the query in the embedding vector space.
RAG's retrieval nature has many benefits; it reduces the input size to process, which reduces the computation during inference~\cite{luo2024bge} and improves the accuracy by eliminating distracting information~\cite{Catav_2023}.

\begin{figure}
  \centering
  \resizebox{\linewidth}{!}{
  \includegraphics[]{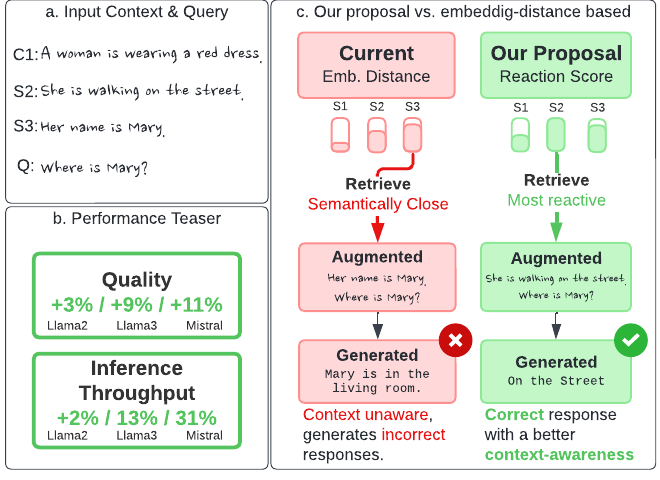}
  }
  \caption{
  YOURA improves the retrieval quality and inference throughput by retrieving only the ``reactive'' sentences.
  }
  \label{fig:teaser}
\end{figure}


A high-quality retrieval is crucial for RAG but often suffers from lower accuracy because (1) the common words in both the query and text chunk put the semantic distance closer and (2) the semantically identical, but alphabetically different words distance the query and text chunk in the embedding vector space.
For example, the similarity distance between a query (``Where is Mary?'') and a sentence (``Her name is Mary'') may be closer to a sentence (``She is at home'') resulting in suboptimal retrieval.
\footnote{Measured with Sentence-Transformer ``All-MiniLM-L6-V2'', scipy.spatial.distance.}


\begin{figure*}
  \centering
  \resizebox{\textwidth}{!}{
  \includegraphics[]{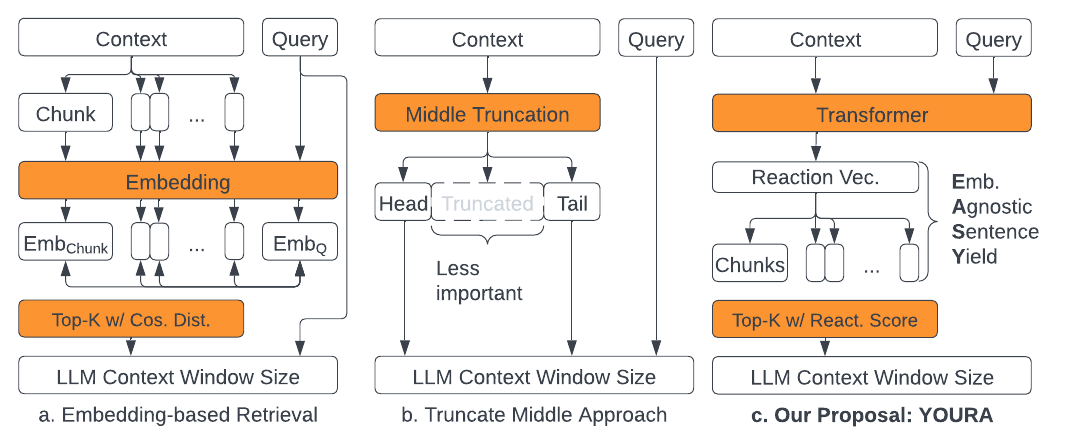}
  }
  \caption{Comparison of a typical embedding-based retriever (cosine similarity distance), truncation-based approach, and our attention-based retriever (reaction score). In our evaluation, we show that reaction score is a better retrieval heuristic in terms of quality and performance. Unlike the embedding-based approach, which splits the raw string context into chunks (e.g., sentences) before retrieval, YOURA splits the token-indexed vector, \textit{reaction-vector} in (c), into chunks. Such an approach requires mapping the raw sentences to token indices. We propose Embedding Agnostic Sentence Yield (EASY) algorithm for the challenge.
  }
  \label{fig:compare}
\end{figure*}

We propose a fine-tuning-free, attention-based retrieval strategy, which can be applied to various off-the-shelf pre-trained LLMs to achieve improved text generation quality. 
Our \emph{key insight} is that the attention in a pre-trained model already contains information about which tokens are important for the given query.
Figure~\ref{fig:compare} illustrates the difference between the embedding-based retrieval and our proposal; (a) is a typical embedding-based retrieval where embedding vectors for both the query and context chunk list are generated through an embedding layer.
Figure~\ref{fig:compare} (b) is another simple yet effective alternative to the embedding distance-based approach, truncate-middle.
Since the introduction and conclusion of most documents contain the most important information, the truncation approach naively takes out the middle of the context so that the remaining text fits within the LLM's context window size.
Unlike embedding-based retrieval or truncation, we propose an attention-based retrieval.

Our proposed technique, \underline{Y}ou \underline{O}nly \underline{U}se \underline{R}eactive \underline{A}ttention slice (YOURA) Figure~\ref{fig:compare} (c), is a novel attention-based retrieval technique.
Anonlogically speaking, we use the attention vector generated without query as the baseline and measure how each token ``reacts'' to the query.
We refer to the ``reactiveness'' as \textit{Reaction Vector}, an absolute difference in attention scores calculated with and without the query.
The per-sentence average of the reaction vector called \textit{reaction score}, is a good ranking metric when retrieving relevant sentences, as we show in our evaluation.

Properly slicing the reaction vector into sentences is crucial because YOURA splits the token sequence into chunks (as opposed to prior works, which split the raw text into chunks).
Such a difference sparks a new challenge: mapping sentences to a token sequence.
Merely associating specific token values (e.g., punctuation token) to a sentence boundary is insufficient because different tokenizer uses different vocabulary.
For example, a tokenizer may encode \colorbox{gray!15}{.The} (a sentence boundary with a missing white space) as two tokens: [\colorbox{gray!15}{.T}, \colorbox{gray!15}{he}].
In such case, \colorbox{gray!15}{.T} token becomes part of the two consecutive sentences.
Treating the token before or after \colorbox{gray!15}{.T} token as a sentence boundary would easily garble the remaining sentence splits.
For this challenge, we propose the Embedding-Agnostic Sentence Yield (EASY) --- a best-effort token wiggling algorithm --- that takes a list of raw sentences and a single token sequence as input and returns token indices for each sentence.

We evaluate (1) how much YOURA improves the answer quality for the 6 LongBench QA datasets~\cite{bai2023longbench}, (2) how YOURA reduces the inferencing computation cost by retrieving fewer, yet sufficient tokens for proper question-answering, and (3) how well our EASY algorithm maps raw texts to the token-indices.
Without any pre-training or additional embeddings, YOURA shows a higher retrieval ratio (context token count / retrieved token count) for open-source pre-trained models (Llama2, Llama3, Mistral) by up to $\times 1.3$ and still achieves a higher answer quality than prior embedding-based retrieval approaches.
With the smaller number of retrieved tokens, YOURA improves up to 30\% vLLM serving throughput.
Our EASY algorithm shows an average of $93.7$\% sentence-to-token-index mapping accuracy.

We contribute the following.
\begin{itemize}
    \item We propose You Only Use the Reactive Attention slice (YOURA) algorithm, a fine-tuning-free, attention-based retrieval algorithm. YOURA improves QA quality by up to 10\% while improving inference throughput by up to 30\%\footnote{The code has been open-sourced and is available at \href{https://github.com/yjsoh/youra}{https://github.com/yjsoh/youra}.}.
    \item We propose an Embedding-Agnostic Sentence Yield (EASY) algorithm, which can map individual sentences to their vector slice at 93\% accuracy.
    \item We provide a detailed analysis of YOURA's QA quality improvement on LongBench QA datasets.
\end{itemize}


\section{Related Work}

\paragraph{Long-context Language Models.}
Training LLMs on sequences with a maximum length while still ensuring they infer well on sequences longer than the maximum is challenging. 
Previous studies proposed techniques such as positional interpolation~\cite{chen2023extending,li2023functional}, positional extrapolation~\cite{press2021train}, external emory~\cite{wu2020memformer,martins2021infty}, memory-retrieval augmentation strategies~\cite{mohtashami2023landmark,wang2024augmenting} to push the limit of the context window size to 2 million and even longer~\cite{ding2024longrope,bertsch2024unlimiformer}.
Although these works improved model accuracy on various tasks, the large context window size implies more computational and memory costs.

\paragraph{Retrieval-Augmented Generation.}
Retrieval-Augmented Generation (RAG) retrieves relevant document chunks from the external knowledge base. 
Prior works integrate RAG with language models for question answering with long documents~\cite{stelmakh2022asqa} and in open-domain~\cite{giorgi2022exploring}. Furthermore, language models adopted retrieval at various stages such as pretraining~\cite{wang2023shall,izacard2023atlas,borgeaud2022improving}, fine-tuning~\cite{jiang2022retrievalattentionendtoendlearning,zhang2024raftadaptinglanguagemodel} and inference stage~\cite{khandelwal2019generalization}.

\paragraph{Attention Mechanism.} 
Researchers proposed self-attention alternatives to mitigate its quadratic complexity, which becomes a computational bottleneck for a long context. 
These include sparse attention mechanisms with pre-defined sparsity patterns~\cite{zhu2021long,liu2023deja}, recurrence-based method~\cite{bulatov2022recurrent}, low-rank projection attention~\cite{zhao2024galore}, and memory-based mechanisms~\cite{lou2024sparserfastermoreefficient}. 
However, these approximate methods introduce inductive bias (e.g., pre=defined sparsity) that can fit well for specific domains, but may reduce model quality in general LLM training.

\paragraph{Fine-tuning for Long Context.} 
Recent works~\cite{ding2024longrope,chen2023longlora,peng2023yarn} show that a pre-trained LLM context window can be extended by fine-tuning on longer texts. 
However, fine-tuning approaches still require multiple self-attention computations demanding both high-quality training datasets and computational costs.
Our approach does not require fine-tuning.

\paragraph{Context Compression.}
LongLLMLingua~\cite{jiang2023longllmlingua} is one of the closest approaches to ours; they observed that the small model is capable of identifying the key relevant information, and devised a way to mitigate the ``positional bias'' due to the positional embeddings by shuffling the long context.
Unlike their approach, which relies on a relatively complex perplexity calculation, our approach merely simply takes the mean within the attention matrix. 

\section{You Only Use Reactive Attention (YOURA)}
\label{sec:youra}

\paragraph{Limitation of Embedding Vectors --- Context-Unaware}
The limitation of embedding-based retrieval is that the input texts are the sole variable when generating the embedding vectors.
For example, consider the following two paragraphs each with two sentences: ``Her name is Mary. She is at home." and "Her name is Alice. She is at home.".
In both cases, "She is at home" is embedded into the identical vector, resulting in equal distance to any sentence.

\paragraph{Attention-Based Retrieval --- Context-Aware}
We propose an attention-based heuristic for context-aware retrieval.
At a high level, our proposal is measuring how the query sentence impacts the language model's attention distribution.
We observe that attention values of relevant information change more drastically than irrelevant information.
From this observation, we introduce \textit{reaction vector} to quantify the attention shift caused by the query sentence.

\begin{figure*}[t]
  \centering
  \includegraphics[width=\textwidth]{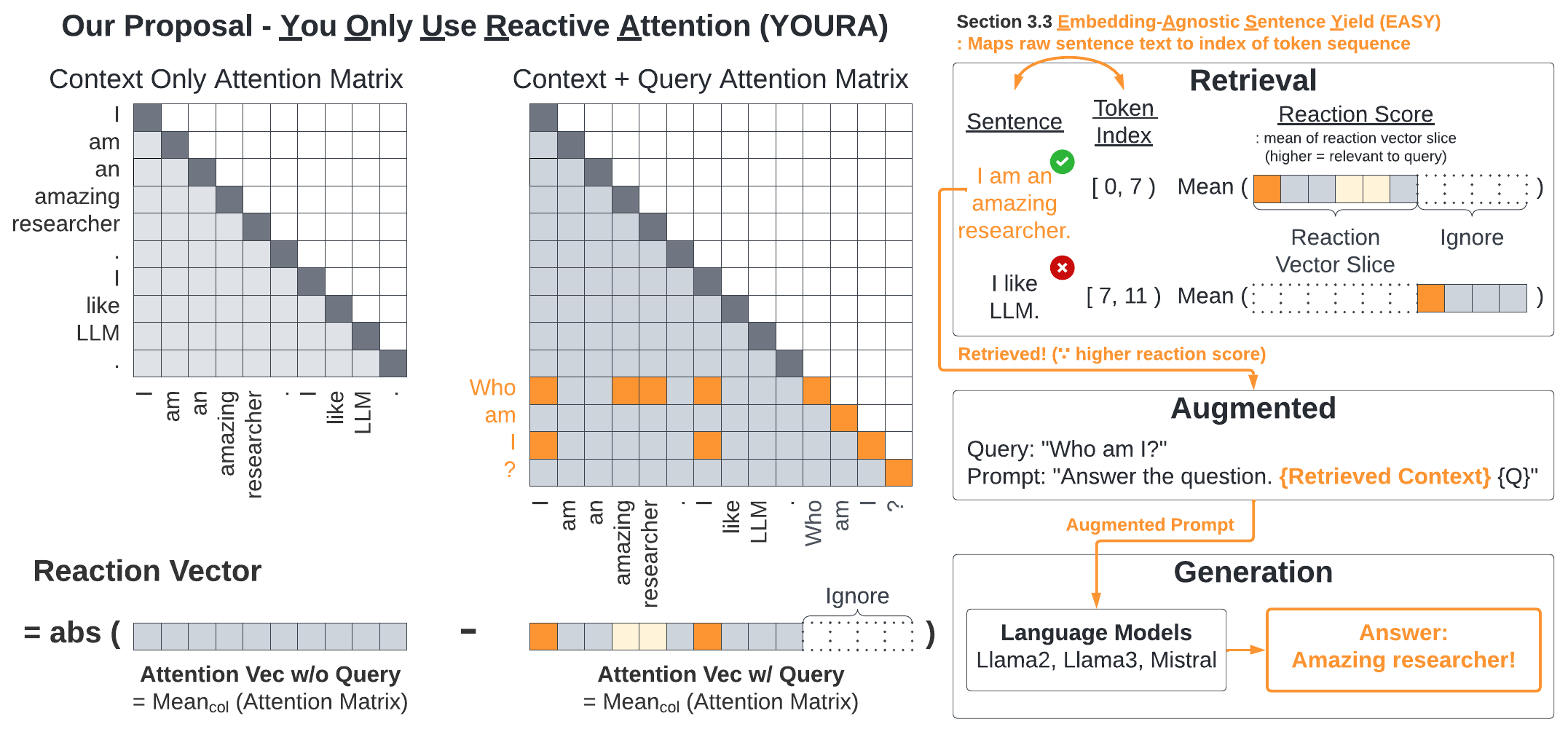}
  \caption{Overview of YOURA and where it is used in the Retrieval Augmented Generation (RAG) with an example (example context: "I am an amazing researcher. I like LLM.", example query: "Who am I?").
  The first step is calculating the \textit{reaction vector}, an absolute difference between the attention vector with and without the query (left side of the figure). 
  The highlighted cells in the attention matrix indicate that the token pair exhibits a relatively high value (e.g., Who vs. I).
  Once the reaction vector has been calculated, each sentence is assigned a \textit{reaction score}, the mean of a corresponding reaction vector slice. 
  To map each sentence to a token sequence, we propose \underline{E}mbedding-\underline{A}gnostic \underline{S}entence \underline{Y}ield (EASY) algorithm (Section~\ref{sec:easy}).
  The retriever passes on the sentences with high reaction scores to the augmentor.
  The pre-trained LLM models generate answers using the augmented text, which includes the task-specific prompt, the retrieved context, and the question.}
  \label{fig:overview}
\end{figure*}

\paragraph{Design Overview}
Figure \ref{fig:overview} illustrates the overview of our proposed design.
The left portion of the figure depicts how a reaction vector is calculated and the right portion depicts how a reaction vector is used for the retrieval and how the language model leverages the retrieved sentences to answer the query.

\subsection{Problem Statement \& Definitions}

\paragraph{Annotations}
We annotate $\mathrm{Z}^{\text{C}} \in \mathbb{R}^{d \times c}$, $\mathrm{Z}^{\text{Q}} \in \mathbb{R}^{d \times q}$ as the continuous representation of the \textbf{context} and \textbf{question}, respectively, and $d$ is the pre-trained model's hidden dimension, $c$ is the context token count and $q$ is the question token count.

\paragraph{Reaction Vector}
We define the \textit{reaction vector} as follows:
\begin{multline}
\text{ReactionVector}(\mathrm{Z}^{\text{C}}, \mathrm{Z}^{\text{Q}}) \\
= \text{abs}\left( \text{AttnVec}(\mathrm{Z}^{\text{C}})
- \text{AttnVec}(\text{concat}(\mathrm{Z}^{\text{C}}, \mathrm{Z}^{\text{Q}}))_{1:c} \right)
\end{multline}
where
\begin{equation}
    \text{AttnVec}(\mathrm{Z}) = \text{Mean}_{col}(\text{AttnMatrix}(Q, K))
\end{equation}

$\text{AttnMatrix}$ is the dense attention function before multiplying the value projection and $Q, K$ are the query and key projection of $Z$, respectively, similar to the description in the ``Attention is all you need'' paper ~\cite{vaswani2023attention}:
\begin{equation}
\begin{aligned}
\text{AttnMatrix}(Q, K) &= \text{softmax}\left(\frac{QK^T}{\sqrt{d}}\right) \\
Q &= \mathrm{Z} \times W^Q \\
K &= \mathrm{Z} \times W^K
\end{aligned}
\end{equation}

$\text{Mean}_{col}$ takes the attention matrix and returns a per-column mean vector (note the subscription $1:c$ to indicate truncation from $1$ to $c$-th column):
\begin{equation}
\begin{aligned}
\text{Mean}_{col}([a_{ij}]_{d \times n})_{1:c} &= \frac{1}{d} \left( \sum_{i=1}^{d}a_{i1}, \sum_{i=1}^{d}a_{i2}, ..., \sum_{i=1}^{d}a_{ic} \right) \\
&= \frac{1}{d} \left( \sum_{i=1}^{d} a_{ij}\right)_{j=1}^c
\end{aligned}
\end{equation}

For multi-head attention, the reaction vector is a mean across heads:
\begin{equation}
\begin{aligned}
\text{Mean}_{col, heads} \left( [a_{ij}^h]_{d \times n} \right)_{1:c} &= \frac{1}{H} \sum_{h=1}^{H} \left( \frac{1}{d} \sum_{i=1}^{d} a_{ij}^h \right)_{j=1}^c \\
&= \frac{1}{H} \left( \sum_{h=1}^{H} \frac{1}{d} \sum_{i=1}^{d} a_{ij}^h \right)_{j=1}^c
\end{aligned}
\end{equation}

\paragraph{Reaction Score}
Given the reaction vector, we define the \textit{reaction score (rs)} as follows for a vector slice represented with left-right index, $[l, r)$:
\begin{multline}
\text{ReactionScore}(\text{ReactionVector}, l, r) \\
= \text{avg}(\text{ReactionVector}[l:r])
\end{multline}

\begin{table*}[t]
    \centering
    \begin{tabular}{l|c|l|r|r|r|c|c}
    \midrule
    Dataset & ID & Source & \multicolumn{3}{c|}{Avg \# Tokens} & Metric & \# Data\\
    &&& Llama2 & Llama3 & Mistral &&\\
    \midrule
    \multicolumn{8}{l}{Single-Document QA}\\
    \midrule
    NarrativeQA & 1-1 & Literature, Film & 35937 & 29777 & 35937 & F1 & 200\\
    Qasper & 1-2 & Science & 5603 & 4922 & 5517& F1 & 200\\
    MultiFieldQA-en & 1-3 & Multi-field & 8046 & 6889 & 7877 & F1 & 150\\
    \midrule
    \multicolumn{8}{l}{Multi-Document QA}\\
    \midrule
    HotpotQA & 2-1 & Wikipedia & 15244 & 12780 & 14890 & F1 & 200\\
    2WikiMultihopQA & 2-2 & Wikipedia & 8381 & 7096 & 8281 & F1 & 200\\
    MuSiQue & 2-3 & Wikipedia & 18459 & 15544 & 18069 & F1 & 200\\
    \midrule
    \end{tabular}
    \caption{Evaluated dataset details.}
    \label{tab:datasets}
\end{table*}
\paragraph{Problem Statement}

We define the retrieval task as finding a set of non-overlapping chunks \(\mathit{S} = \{(l_i, r_i) \mid 0 \leq l_i \leq r_i \leq |\mathrm{Z}^{\text{ctxt}}|, \, r_i \leq l_{i+1} \}\), such that
\begin{multline}
\text{argmax}_{\mathit{S}} \, \text{AnsQuality}(\text{LLMInference}(\text{concat}(\mathit{S}, q)))
\end{multline}
\noindent where:
\begin{itemize}
    \item \(\text{concat}(\mathit{S}, q)\) denotes the concatenation of the chunks in \(\mathit{S}\) with the query \(q\).
    \item \(\text{LLMInference}\) is the function that processes the concatenated chunks and the query using a language model.
    \item \(\text{AnsQuality}\) measures the quality (e.g., F1 Score) of the answer produced by the inference function.
\end{itemize}

\subsection{Embedding-Agnostic Sentence Yield (EASY) Algorithm}
\label{sec:easy}
\paragraph{Challenge}
Given a token sequence, finding the token index of sentence boundary is non-trivial for two reasons: (1) the embedding model for sentence split task (e.g., stanza) is different from the model's embedding, and (2) embedding models with huge dictionary may result in an impossible token sequence to achieve a perfect sentence split.
For example, widely known sentence-splitting algorithms (e.g., Stanza~\cite{qi2020stanza}) use a different dictionary from the LLM models (e.g., Llama3-8B-Instruct).
Thus, the total token count varies across the embedding models (Table~\ref{tab:datasets}).

Another challenge is that some models prefer huge dictionaries resulting in a token sequence with an impossible case to articulate a sentence boundary.
For example, llama-3 treats \colorbox{gray!15}{.T} as a single token.
Although it may be useful in code generation, a missing whitespace between a punctuation period and a subsequent sentence that starts with the word ``The'' would result in an impossible case.
In other words, Stanza would create a token sequence (\colorbox{gray!15}{.} - \colorbox{gray!15}{The}), whereas Llama3 tokenizer would output (\colorbox{gray!15}{.T} - \colorbox{gray!15}{he}).
In such a case, an exact sentence boundary for Llama3 cannot be expressed as a token index.
For these reasons, finding the sentence boundary token index across the embedding model is non-trivial.

\begin{algorithm}[t]
\caption{EASY Algorithm}
\label{alg:easy}
\begin{algorithmic}[1]
\REQUIRE $seq$ - Encoded token sequence
\REQUIRE $TS$ - \underline{T}arget \underline{S}entences generated by NLP models
\REQUIRE $tol \gets 30$ - Max wiggling tries

\STATE $P \gets []$  \COMMENT{Processed sentences}
\STATE $B \gets []$  \COMMENT{Sentence boundary indices}
\STATE $i, m \leftarrow 0$  \COMMENT{Initialize indices}

\WHILE{$P \neq TS$}
    \STATE APPEND $TS[i]$ to $P$
    \STATE $c \leftarrow$ Length(Encode(Concat($P$))) \COMMENT{Candidate}
    \STATE $saved \gets c$
    \STATE $visited \gets \emptyset$
    \WHILE{true}
        \IF{$c \in visited$ or $c > |seq|$ or $|c - saved| > tol$}
            \STATE $c \gets saved$
            \STATE \textbf{break}
        \ENDIF
        \STATE ADD $c$ to $visited$
        \STATE $s \leftarrow$ Decode($seq[m:c]$)
        \IF{$s == TS[i]$}
            \STATE \textbf{break}
        \ELSIF{$s$ is substring of $TS[i]$}
            \STATE $c \leftarrow c + 1$
        \ELSE
            \STATE $c \leftarrow c - 1$
        \ENDIF
    \ENDWHILE
    \STATE APPEND $c$ to $B$
    \STATE $m \leftarrow c$
    \STATE $i \leftarrow i + 1$
\ENDWHILE

\STATE \textbf{return} $B$
\end{algorithmic}
\end{algorithm}

\paragraph{Embedding-Agnostic Sentence Yield (EASY)}
To resolve the problem, we propose the Embedding-Agnostic Sentence Yield (EASY) algorithm.
The input to the algorithm is a single token sequence and a list of sentence strings acquired from the LLM tokenizer and the sentence splitting NLP models (e.g., Stanza~\cite{qi2020stanza}), respectively.

The EASY algorithm is a repetition of guessing the sentence boundary and making best efforts adjustments.
First, EASY estimates a candidate index as the encoded length of sentences that are already processed (Algorithm~\ref{alg:easy} line 7).
EASY decodes the token subsequence between the \textbf{most-recently confirmed boundary index} ($m$ in Algorithm~\ref{alg:easy}) and the \textbf{current candidate} and compares it against the target sentence (Algorithm~\ref{alg:easy} line 16).
Based on the comparison outcome (\texttt{MATCH}, \texttt{INCLUDED}, \texttt{NO-MATCH}), EASY wiggles (+1 or -1) the candidate index until the decoded subsequence matches the target or a termination condition is met (lines 17-23).
Lastly, EASY returns the list of integers, where each value is a token index corresponding to a sentence boundary.
If for some reason, the best effort fails, EASY restores the initial candidate index and appends to the boundary list.
This retains the one-to-one mapping between the target sentence and the boundary token index.

\subsection{Retrieval using Reaction Score --- End-to-end}
The retrieval algorithm starts with calculating the reaction vector.
If the input context is longer than the model's context window size, we calculate the attention vector for each chunk of context window size and concatenate them before returning.
At this point, we have the reaction vector whose length equals the input context's token sequence.

To slice the reaction vector for each sentence, we apply the Embedding-Agnostic Sentence Yield (EASY) algorithm (Section \ref{sec:easy}).
Then the retriever slices the reaction vector and calculates the geometric mean of the reaction vector slice using the output of the EASY algorithm, a list of sentence boundary indices.
At this point, we have a list of sentences and their associated reaction scores, which will be used as a retrieval heuristic.

For the actual retrieval, we greedily add sentences starting with the highest reaction score until the retrieval token budget is depleted or the sentence count threshold is reached (80\% of total sentence count).
The retrieved sentences are reordered according to their original position as the final cleanup step.
\section{Experiments}

We evaluate how the YOURA improves the answering performance of 5 pre-trained models on LongBench~\cite{bai2023longbench} single-doc and multi-doc QA datasets.

\paragraph{Datasets} 
We evaluate YOURA on the single and multi-document QA (Table~\ref{tab:datasets}).

\paragraph{Models}
We used three different open-sourced LLM models: LLama2-7B, Llama3-8B, and Mistralv0.2-7B, where each model has a context window size of 4K, 8K, and 32K, respectively.

\paragraph{Inference Setup}
For inference, we used the vLLM~\cite{kwon2023efficient}.
We used the LongBench~\cite{bai2023longbench} code to evaluate the generated answers.

\paragraph{Machine Setup}
We measured the vLLM~\cite{kwon2023efficient} inference performance by slightly modifying the vLLM's throughput benchmark (v0.5.3) code on a single NVIDIA H100-80G GPU.
We used the default vLLM settings except for changing the \colorbox{gray!15}{ignore\_eos} to \colorbox{gray!15}{false}.

\paragraph{Experiment Setup}
We compare the following scenarios to evaluate how the YOURA improves the output quality or performance. For each setup, as done with the LongBench~\cite{bai2023longbench}, we equalize the retrieval budget to 3500, 7500, and 31500 for models with 4K, 8K, and 32K context window sizes, respectively.

\begin{itemize}
    \item \textbf{Model without context} No context is passed to the LLM.
    \item \textbf{with BM25 Retrieval} \colorbox{gray!15}{B$x$\_$y$} indicates chunking at roughly $x$ tokens and retrieving the top $y$ chunks.
    \item \textbf{with Embedding Retrieval} \colorbox{gray!15}{E$x$\_$y$} indicates chunking at roughly $x$ tokens and retrieving the top $y$ chunks. We measure the cosine similarity between each chunk's embedding vector generated by the ``paraphrase-MiniLM-L6-v2'' model\footnote{https://huggingface.co/sentence-transformers/paraphrase-MiniLM-L6-v2}.
    \item \textbf{with Truncation} We use LongBench's prompt along with its context truncation approach: the whole context, if it fits within the pre-trained model's context window size, or the concatenation of the context head and tail with an equal context window budget, otherwise.
    \item \textbf{with YOURA} Our proposed system as described in Section~\ref{sec:youra}.
\end{itemize}

\begin{table}[t!]
    \centering
    \begin{tabular}{l|c|r|r|r}
    \midrule
        Setup & Retr. Ratio & SDOC & MDoc & Overall\\
               & Avg       & Avg  & Avg  & Avg\\
        \midrule
        \midrule
        LLaMA2 & INF    & 13.65 & 19.30 & 16.47\\
      w/ B500\_7 & 3.28 & 22.82 & 20.17 & 21.49\\
      w/ B50\_70 & 3.28 & 23.86 & 23.25 & 23.56\\
      w/ E500\_7 & 3.28 & 17.13 & 19.57 & 18.35\\
      w/ E50\_70 & 3.28 & 24.15 & 22.87 & 23.51\\
      w/ Trunc. & 3.28 & \textbf{25.74} & 22.34 & 24.04\\
      w/ YOURA & 3.39 & 23.22 & \textbf{25.17} & \textbf{24.20} \\
        \midrule \midrule

        LLaMA3  & INF  & 11.59 & 23.16 & 17.37 \\
        w/ B500\_15 & 1.57 & 34.07 & 33.27 & 33.73 \\
        w/ B50\_150  & 1.57 & 32.92 & 35.83 & 34.17 \\
        w/ E500\_15 & 1.57 & 33.33 & 30.82 & 32.08 \\
        w/ E50\_150 & 1.57 & 31.50 & 35.70 & 33.60 \\
        w/ Trunc. & 1.57 & \textbf{36.83} & 34.72 & 35.93 \\
        w/ YOURA & 1.76 & 36.57 & \textbf{36.44} & \textbf{36.50}\\
        \midrule \midrule

        Mistralv0.2 & INF    &  7.02 & 11.47 &  9.25 \\
          w/ B500\_63 & 1.04 & 32.29 & 24.66 & 29.02 \\
          w/ B50\_630 & 1.04 & 29.14 & 23.28 & 26.63 \\
          w/ E500\_63 & 1.04 & 26.79 & 22.06 & 24.76 \\
          w/ E50\_630 & 1.04 & 29.14 & 23.28 & 26.63 \\
          w/ Trunc.   & 1.04 & 32.65 & \textbf{26.62} & \textbf{29.64} \\
        w/ YOURA & 1.33 & \textbf{33.14} & 25.84 & 29.49 \\

        \midrule

    \end{tabular}
    \caption{Answer quality comparison for LongBench's single-document and multi-document QA datasets. YOURA shows better quality with a higher retrieval ratio --- i.e., retrieved fewer tokens. All retrieval scenarios use the maximal budget (4K, 8K, and 32K tokens depending on the model) except for without-context setup and ``YOURA''. The retrieval ratio (second column) is the average of the total token count divided by the retrieved token count.}
    \label{tab:result}
\end{table}

\subsection{Longbench QA Result}

\textbf{How does YOURA impact the output quality?}
From Table \ref{tab:result}, which shows the answer quality using three different open-sourced models (Llama2, Llama3, and Mistralv0.2), we make the following observations. 
(1) We observe that YOURA showed better overall answer quality than all of the retrieval approaches (BM25 and embedding-based, except for Llama2 w/ B50\_70) at a higher retrieval ratio --- i.e., retrieving less information from the long context.
(2) When compared to the truncate-middle approach, YOURA shows better overall quality for Llama2 and Llama3, and slightly less (-0.15) for Mistral.

(2-1) When broken down into dataset types, we observe that YOURA is better at multi-document QA than single-document QA for Llama2 and Llama3.
We claim that Trunc. truncates the key information to correctly answer multi-document QAs, but the head and tail of a single document include key information in many cases.
As for YOURA, the key information is retrieved properly for both the single-doc and multi-doc QAs resulting in a comparable answer quality.
For YOURA's lower single-document score, we conjecture that the retrieval budget (4k, and 8k for Llama2 and Llama3, respectively) is insufficient to retrieve all of the relevant information.
As a supporting fact, the quality difference between Trunc. and YOURA decreases significantly from -2.52 to -0.26.
Experiment results for a pre-trained model with a 16k context window size would be useful for proving/disproving the conjecture but no solid 16k model is available at the time of the publication.

(2-2) For Mistral, which has a large context window size of 32k, truncation rarely happens (as shown in the small retrieval ratio).
Without truncation, distractful information is included as part of the augmented context resulting in worse single-document answer quality than YOURA.
\footnote{The impact of distractful information on output quality is observed in other work~\cite{Catav_2023} as well.}
Our explanation for multi-document QA using Mistral, where YOURA showed a slightly worse quality (-0.78), is similar to the explanation in (2-1); YOURA's relatively high retrieval ratio resulted in leaving out some key information.

From the observations, we conclude that the balance between the pre-trained model's context window size and the retrieval ratio is important for long context QA.

\begin{table}
    \centering
    \begin{tabular}{l|c|c|c|c}
    \midrule
    Mod & Retrieval & SDOC & MDOC & Overall Avg\\
    \midrule
    L2 & Trunc. & 5.25 & 5.76 & 5.50 \\
    L2 & YOURA  & \textbf{5.43} & \textbf{5.78} & \textbf{5.60 (+2\%)} \\
    \midrule
    
    L3 & Trunc. & 3.23 & 2.79 & 3.01 \\
    L3 & YOURA & \textbf{3.76} & \textbf{3.03} & \textbf{3.39 (+13\%)} \\
    \midrule
    
    M & Trunc. & 1.97 & 1.36 & 1.66\\ 
    M & YOURA  & \textbf{2.54} & \textbf{1.82} & \textbf{2.18 (+31\%)} \\
    \midrule
    \end{tabular}
    \caption{vLLM Serving performance (request per second, higher the better) for LongBench's single-document and multi-document QA datasets with respect to Truncation and YOURA. YOURA retrieved fewer tokens resulting in up to 30\% higher inference throughput.}
    \label{tab:vllm}
\end{table}

\textbf{How does YOURA's higher retrieval ratio improve inference performance?}
Table~\ref{tab:vllm} shows the average vllm inference throughput (requests per second) assuming offline truncation/retrieval.
We make the following observations. 
First, YOURA improves the overall inference throughput, especially for models with larger context window sizes.
This is because the relative retrieval ratio is larger for models with larger context window sizes.
Second, the retrieval ratio is a good relative throughput estimator. 
The relative throughput improvements (+2\%, +13\%, +31\%) are on par with the relative retrieval ratio (+3\%, +12\%, +28\%).
We conclude that the performance of LLM serving platforms such as vLLM is sensitive to the retrieved context size and underscores the importance of good information retrieval.

\subsection{EASY Sentence Splitter Accuracy}

\paragraph{How well can the EASY algorithm split tokens into sentences?}
Table \ref{tab:easy_result} shows the quality of our EASY sentence splitting algorithm.
We used Stanza~\cite{qi2020stanza}, an open-sourced NLP toolkit, to split raw text into sentences. 
Then we ran the EASY algorithm for each tokenizer model to identify the sentence boundary within a token sequence.
After decoding each token segment (delimited by the sentence boundary index), it is a match if the Stanza's sentence (namely \textit{target sentence}) matches exactly after the cleanup (e.g., trailing whitespace).
We report the average match rate for each model and dataset pair.

We make the following observations.
Overall the EASY algorithm can identify the sentence boundary for models with varying dictionary sizes: Llama2, Llama3, and Mistralv0.2 tokenizers with 32000, 128256, and 32000 words, respectively.
Llama2 and Mistral showed almost identical match ratios, as both models' tokenizers are based on similar vocabulary.
Llama3 uses roughly $\times 4$ more vocabulary resulting in fewer total token counts as shown in Table~\ref{tab:datasets} (e.g., \colorbox{gray!15}{.T} into a single token).
Due to the rich vocabulary, a typo easily blurred sentence boundaries causing many imperfect matches and thus the smallest overall match quality as in the 2-1, 2-2, and 2-3 datasets.
For 1-1, 1-2, and 1-3, datasets, the rich vocabulary worked in favor.
Many atypical char sequences (e.g., $<$b$>$, math formula) were present in these datasets, which were properly represented with tokens using Llama3.

\begin{table}[]
    \centering
    \begin{tabular}{cccccccc}
        \midrule
         Mod & 1-1 & 1-2 & 1-3 & 2-1 & 2-2 & 2-3 & Avg\\
         \midrule
         \midrule
         L2 & 94.1 & 83.1 & 93.5 & 98.4 & 98.8 & 97.8 & 94.3 \\
         L3 & 99.9 & 99.6 & 96.5 & 86.5 & 86.9 & 85.6 & 92.5\\
         M  & 94.0 & 83.1 & 93.5 & 98.4 & 98.8 & 97.8 & 94.3\\

         \midrule
         Avg & 96.0 & 88.6 & 94.5 & 94.4 & 94.8 & 93.7 & 93.7\\
         \midrule
         \multicolumn{8}{r}{L2: Llama2-7B, L3: Llama3-8B, M: Mistralv0.2}\\
    \end{tabular}
    \caption{Evaluation of EASY algorithm's average sentence match rate (exact match / total \# of sentences) for each model and dataset (Table~\ref{tab:datasets}) pairs.}
    \label{tab:easy_result}
\end{table}

\paragraph{How resilient is the EASY algorithm from edge cases?}
To quantify the resilience despite typos and overly representative dictionaries, we measured the Levenshtein distance between the target sentence and the sentence yielded by our EASY algorithm (Table~\ref{tab:easy_lev}).
On average, one needs less than 3 character edits (insert, remove, replace) to the EASY output sentences to match the target sentences.

The NZ row in Table~\ref{tab:easy_lev} is the average Levenshtein distance of unmatched sentences.
From these rows, we conclude that the EASY algorithm is resilient to incorrect split and its cascading effect.

\begin{table}[t!]
    \centering
    \begin{tabular}{lccccccc}
    \midrule
     Mod & 1-1 & 1-2 & 1-3 & 2-1 & 2-2 & 2-3 & Avg\\
     \midrule
     \midrule
     L2 & 3.26 & 3.48 & 2.54 & 2.61 & 2.36 & 2.67 & 2.82\\
     NZ & 5.40 & 4.80 & 4.35 & 4.22 & 4.13 & 4.31 & 4.54\\
       \midrule
     L3 & 1.71 & 3.45 & 2.77 & 3.16 & 2.80 & 3.23 & 2.85\\
     NZ & 3.70 & 4.76 & 4.52 & 4.53 & 4.29 & 4.64 & 4.41\\
       \midrule
     M & 3.54 & 3.52 & 2.56 & 2.65 & 2.38 & 2.71 & 2.89\\
     NZ & 5.52 & 4.85 & 4.40 & 4.29 & 4.17 & 4.38 & 4.60\\
    \midrule
     \multicolumn{8}{r}{L2: Llama2-7B, L3: Llama3-8B, M: Mistralv0.2}\\
     \multicolumn{8}{r}{NZ* (Non-Zero): average of non-zero distances.}\\
    \end{tabular}
    \caption{Average Levenshtein distance between the target sentence and EASY algorithm's sentence split. 
    A smaller number indicates fewer character edits (insert, remove, replace) to transform the EASY output sentence to the target sentence, and thus a better quality split.}
    \label{tab:easy_lev}
\end{table}

\section{Conclusion}

We propose a simple, yet effective attention-based retrieval algorithm.
We show that attention vectors are useful retrieval and propose novel retrieval heuristics called reaction scores.
Our fine-tuning approach improves the output quality or the inference performance of off-the-shelf models for long-context QA tasks.


\newpage
\bibliography{paper}
\bibliographystyle{mlsys2024}

\clearpage
\appendix
\section{Detailed Results}
Table~\ref{tab:resultfull}, Table~\ref{tab:vllmfull} are the detailed version of Table~\ref{tab:result} and Table~\ref{tab:vllm}, respectively.

\begin{table*}[t!]
    \centering
    \begin{tabular}{l|c||r|r|r|r||r|r|r|r||r}
    \midrule
        Setup & Retr. Rate & \multicolumn{3}{c|}{Single-Document QA} & SDoc & \multicolumn{3}{c|}{Multi-Document QA} & MDoc & Overall\\
         & Avg & 1-1 & 1-2 & 1-3 & Avg & 2-1 & 2-2 & 2-3 & Avg & Avg\\
        \midrule
        \midrule
        LLaMA2 & INF & 8.89 & 13.87 & 18.19 & 13.65 & 23.05 & 25.67 & 9.17 & 19.30 & 16.47\\
      w/ B500\_7 & 3.28 &  12.59 & 21.44 & 34.42 & 22.82 & 25.92 & 27.47 & 7.11 & 20.17 & 21.49\\
      w/ B50\_70 & 3.28 & 16.37 & 21.51 & 33.71 & 23.86 & 27.48 & 28.62 & 13.64 & 23.25 & 23.56\\
      w/ E500\_7 & 3.28 &  12.85 & 16.08 & 22.47 & 17.13 & 25.51 & 24.84 & 8.35 & 19.57 & 18.35\\
      w/ E50\_70 & 3.28 & 17.15 & 20.12 & 35.18 & 24.15 & 28.80 & 27.15 & 12.66 & 22.87 & 23.51\\
      w/ Trunc. & 3.28 & \textbf{18.84} & 21.71 & \textbf{36.68} & \textbf{25.74} & 27.55 & 31.08 & 8.40 & 22.34 & 24.04\\
      w/ YOURA & 3.39 & 15.48 & \textbf{23.41} & 30.77 & 23.22 & \textbf{29.55} & \textbf{31.11} & \textbf{14.86} & \textbf{25.17} & \textbf{24.20} \\
        \midrule \midrule

        LLaMA3 & INF &  9.62 & 14.38 & 10.78 & 11.59 & 28.43 & 30.46 & 10.60 & 23.16 & 17.37\\
        w/ B500 & 1.57 & 18.27 & 40.21 & 43.72 & 34.07 & 41.75 & 41.18 & 16.88 & 33.27 & 33.73 \\
        w/ B50  & 1.57 & 19.86 & 36.51 & 42.39 & 32.92 & 47.04 & 40.58 & 19.86 & 35.83 & 34.17 \\
        w/ E500 & 1.57 & 16.38 & 39.85 & 43.76 & 33.33 & 42.28 & 34.93 & 15.25 & 30.82 & 32.08 \\
        w/ E50  & 1.57 & 19.05 & 35.44 & 40.02 & 31.50 & 48.05 & \textbf{36.85} & 22.19 & 35.70 & 33.60 \\
        w/ Trunc. & 1.57 & \textbf{21.54} & \textbf{44.21} & 44.75         & \textbf{36.83} & 46.43 & 36.23             & 21.50          & 34.72 & 35.93 \\
        w/ YOURA & 1.76 & 21.47 & 42.39 & \textbf{45.85} & 36.57 & \textbf{50.78} & 34.24 & \textbf{24.29} & \textbf{36.44} & \textbf{36.50}\\
        \midrule \midrule

        Mistralv0.2 & INF & 7.24 & 7.69 & 6.13 & 7.02 & 16.41 & 11.87 & 6.13 & 11.47 & 9.25 \\
          w/ B500\_63 & 1.04 & 20.74 & 29.22 & 46.91 & 32.29 & 34.26 & 22.26 & 17.47 & 24.66 & 29.02 \\
          w/ B50\_630 & 1.04 & 18.84 & 25.08 & 43.50 & 29.14 & 32.85 & 18.66 & 18.32 & 23.28 & 26.63 \\
          w/ E500\_63 & 1.04 & 14.02 & 24.14 & 42.22 & 26.79 & 32.81 & 17.55 & 15.82 & 22.06 & 24.76 \\
          w/ E50\_630 & 1.04 & 18.84 & 25.08 & 43.50 & 29.14 & 32.85 & 18.66 & 18.32 & 23.28 & 26.63 \\
          w/ Trunc. & 1.04 & 20.80 & 29.23 &  \textbf{47.92} & 32.65 & \textbf{37.28} & \textbf{21.72} & \textbf{20.86} & \textbf{26.62} & \textbf{29.64} \\
          w/ YOURA & 1.33 & \textbf{22.37} & \textbf{30.17} & 46.88 & \textbf{33.14} & 36.11 & 21.45 & 19.95 & 25.84 & 29.49 \\

        \midrule
        \multicolumn{11}{r}{The \textbf{bold score} indicates the best score for each pre-trained model.}

    \end{tabular}
    \caption{Detailed experiment results for LongBench's single-document and multi-document QA datasets. YOURA shows better quality with a higher retrieval ratio --- i.e., retrieved fewer tokens. All retrieval scenarios use the maximal budget (4K, 8K, and 32K tokens depending on model) except for without-context setup and ``YOURA''. Retrieval ratio (second column) is an average of the total token count divided by the retrieved token count.}
    \label{tab:resultfull}
\end{table*}

\begin{table*}[t!]
    \centering
    \begin{tabular}{l|c|c|c|c|c|c|c|c}
    \midrule
    Model & Retrieval & \multicolumn{3}{c|}{Single-Document QA}  & \multicolumn{3}{c|}{Multi-Document QA}  & Overall Avg\\
         & Method & 1-1 & 1-2 & 1-3 & 2-1 & 2-2 & 2-3  & Req./Sec \\
        \midrule

    LLaMA2 & Trunc. & 5.58 & 4.68 & 5.50 & \textbf{5.71} & 5.87 & \textbf{5.68} & 5.50 \\
    LLaMA2 & YOURA & \textbf{5.70} & \textbf{4.88} & \textbf{5.72} & 5.70 & \textbf{6.00} & 5.63 & \textbf{5.60 (+2\%)} \\
    \midrule
    
    LLaMA3 & Trunc. & 2.50 & 3.95 & 3.23 & 2.63 & 3.23 & 2.52 & 3.01 \\
    LLaMA3 & YOURA & \textbf{2.60} & \textbf{4.87} & \textbf{3.82} & \textbf{2.72} & \textbf{3.83} & \textbf{2.53} & \textbf{3.39 (+13\%)} \\
    \midrule
    
    Mistralv0.2 & Trunc. & 0.55 & 3.17 & 2.19 & 1.08 & 2.14 & 0.86 & 1.66 \\ 
    Mistralv0.2 & YOURA & \textbf{0.64} & \textbf{4.02} & \textbf{2.97} & \textbf{1.45} & \textbf{2.85} & \textbf{1.15} & \textbf{2.18 (+31\%)} \\
    \midrule
    \end{tabular}
    \caption{vLLM Serving performance (request per second, higher the better) for LongBench's single-document and multi-document QA datasets with respect to Truncation and YOURA. YOURA retrieved fewer tokens resulting in upto 30\% higher inference throughput.}
    \label{tab:vllmfull}
\end{table*}


\end{document}